\documentclass[runningheads]{llncs}
\usepackage[T1]{fontenc}
\usepackage{float}
\usepackage{cite}
\usepackage{hyperref}
\usepackage{subcaption}
\usepackage[ruled,vlined,linesnumbered]{algorithm2e}

\usepackage{graphicx}
\usepackage{amsmath}
\usepackage{amssymb}

\begin{document}
\title{Knowledge-Graph Based RAG System Evaluation Framework}
%
%
\author{Sicheng Dong\orcidID{0009-0006-1347-6421} \and
Vahid Zolfaghari\orcidID{0009-0004-0039-6014} \and
Nenad Petrović\orcidID{0000-0003-2264-7369}
\and Alois Knoll\orcidID{0000-0003-4840-076X}}

\authorrunning{S. Dong et al.}
%
\institute{
Technical University of Munich, Robotics, Artificial Intelligence and Embedded Systems, Munich, Germany\\
\email{\{sicheng.dong, v.zolfaghari, nenad.petrovic, k\}@tum.de}
\url{https://www.ce.cit.tum.de/en/air/home/}
}

\maketitle              
\begin{abstract}
Large language models (LLMs) has become a significant research focus and is utilized in various fields, such as text generation and dialog systems. One of the most essential applications of LLM is Retrieval Augmented Generation (RAG), which greatly enhances generated content's reliability and relevance. However, evaluating RAG systems remains a challenging task. Traditional evaluation metrics struggle to effectively capture the key features of modern LLM-generated content that often exhibits high fluency and naturalness. Inspired by the RAGAS tool, a well-known RAG evaluation framework, we extended this framework into a KG-based evaluation paradigm, enabling multi-hop reasoning and semantic community clustering to derive more comprehensive scoring metrics. By incorporating these comprehensive evaluation criteria, we gain a deeper understanding of RAG systems and a more nuanced perspective on their performance. To validate the effectiveness of our approach, we compare its performance with RAGAS scores and construct a human-annotated subset to assess the correlation between human judgments and automated metrics. In addition, we conduct targeted experiments to demonstrate that our KG-based evaluation method is more sensitive to subtle semantic differences in generated outputs. Finally, we discuss the key challenges in evaluating RAG systems and highlight potential directions for future research.

\end{abstract}
\section{Introduction}
Large Language Models (LLMs) are one of the hottest research topics in artificial intelligence today, and they have proven to be extremely powerful in a variety of fields, including healthcare and education.\cite{b1}\cite{b2} Despite their strong performance, LLMs nevertheless have a number of serious drawbacks. For example, they frequently lack the knowledge required to respond to domain-specific queries.\cite{b3} Furthermore, LLM databases eventually become out of date and cannot address today's issues.\cite{b4}\\
Researchers have taken two primary approaches to solving these issues: \textbf{fine-tuning the model using domain-specific data} and \textbf{connecting the model to additional external information sources}.\cite{b5} Although fine-tuning is a straightforward and effective approach, it has some obvious drawbacks, such as the scarcity of high-quality domain data and the high computational cost of the training process.\cite{b6} As a result, the second strategy, known as the \textbf{Retrieval Augmented Generation (RAG) system}, is increasingly being used in research. By accessing external data sources, this approach can search for domain-specific data in real time without the need for extensive training.\cite{b7} In addition, RAG is also regarded an effective structure to solve the problem of the system to generate inaccurate or misleading information (hallucination).\cite{b8}\\
A RAG system consists of two key components: a \textbf{retriever} and a \textbf{generator}. The retriever will fetch relevant information based on the given input, and the generator then utilizes the information from the retriever to produce the final output.\cite{b9}
Although baseline RAG has demonstrated strong information retrieval capabilities in certain tasks, it still faces several key challenges, particularly its limited ability to integrate multiple information sources. When answering a question requires synthesizing information from multiple, distinct sources, baseline RAG often struggles to effectively combine these pieces into a coherent and accurate response.\cite{b11}\\
To address this issue, recent research has explored the integration of \textbf{Knowledge Graphs (KG)} \cite{b29} with RAG, leading to the development of the \textbf{GraphRAG} architecture.\cite{b11} This approach leverages LLMs for entity recognition and relation extraction to construct KGs, which will then be integrated with graph machine learning metrics to capture a more completed structure of the retrieved information, resulting in high response quality and accuracy.\\
Building on this foundation, the latest research proposed \textbf{lightRAG}, another KG-based RAG system.\cite{b12} It leverages KG to improve retriever capability and optimizes the overall system architecture to provide a more efficient and lighter generation of response.\\
However, when deploying such systems in real-world applications, it becomes critical to understand the reliability and effectiveness of RAG systems. The recently widely adopted evaluation framework, RAGAS, leverages large language models and techniques such as atomic facts to provide a more comprehensive assessment. Although atomic facts are very effective, they still face challenges when dealing with complex documents or when finer-grained evaluation is required. Therefore, we use a knowledge graph here to enhance the evaluation capability in this aspect.
In this experiment, inspired by \textbf{RAGAS}, we extended this to a KG-based approach, aiming to provide a more precise  evaluation system capable of handling complex, multi-fact relationships.

\subsection{Research Questions}

The Research Questions (RQs) in this work are:

\begin{itemize}
    \item \textbf{RQ1:} Can KG-based metrics improve over RAGAS in factuality/faithfulness evaluation? 
    \item \textbf{RQ2:} How well do KG-based metrics correlate with human judgement?
\end{itemize}

The contributions of our work are as follows:

\begin{itemize}
    \item For \textbf{RQ1}, we introduce a KG-based evaluation framework for RAG systems, extending atomic-level assessment principles inspired by RAGAS. By explicitly modeling factual units and their relationships, our approach achieves more fine-grained and faithful evaluations of factual accuracy and content coverage compared to existing baselines.
    \item For \textbf{RQ2}, we calculated correlation \cite{b32}\cite{b26} and sensitivity experiments between our KG-based scores and RAGAS scores, human annotations across multiple metrics. Results show moderate to strong alignment. Sensitivity experienments reveal that, when comparing questions with extreme situations (totally wrong or totally correct answer), the KG-based method outperforms RAGAS and correlates more closely with human judgments, demonstrating its strength in capturing factual alignment and semantic consistency.
\end{itemize}

\section{Related Work}
RAG systems have attracted widespread attention across various fields, as researchers use them to enhance models’ ability to leverage external knowledge. However, when it comes to dynamic knowledge and other complex structures, evaluating these systems has remained a major research challenge.
The whole evaluation process not only includes assessing the quality of the final generated output but also analyzing the retriever's ability to fetch relevant information and examine the interaction between the retriever and generator components.Traditional evaluation methods, such as \textbf{word-overlap-based metrics} (e.g., BLEU\cite{b13}, ROUGE\cite{b14}) or \textbf{pre-trained model-based methods} (e.g., BERTScore\cite{b15}), struggle to effectively capture the semantic richness of modern LLM-generated text and give a perfect evaluation.\\
Therefore, researchers have begun to focus on LLMs as evaluators for assessing RAG systems. For example, Li et al. (2025) define scoring bias and illustrate how perturbations in prompts or answer templates affect judgments.\cite{b34} Shi et al. (2024) specifically study position bias in pairwise comparisons conducted by LLM judges.\cite{b35}
Moreover, Li et al. (2025) show that LLM judges are less stable  when encountering adversarial manipulations and prompt sensitivities.\cite{b36} Compared to traditional evaluation methods, this kind of LLM-driven approach demonstrates great advantages in both efficiency and accuracy since most of the work can be done by LLMs themselves, reducing manual intervention and enhancing sensitivity to linguistic nuances.\cite{b17}\\
Several well-known evaluation frameworks, such as \textbf{RAGAS} \cite{b16}, have already achieved significant progress in this field. These frameworks implement diverse evaluation metrics. Besides traditional metrics, it also leverages LLMs as evaluators to systematically assess RAG systems. The framework defines a wide range of metrics, some of which are outlined below \cite{b16}:\\
\textbf{Factual Correctness} compares how factually accurate the generated response is compared with the reference.\\
\textbf{Faithfulness} measures the consistency between a response and the retrieved context.\\
\textbf{Answer Relevancy} evaluates how relevant the generated response is to the user input.\\
\textbf{Context Relevancy} evaluates how pertinent the retrieved context is to the user input.\\
Among the methods used in RAGAS are techniques such as splitting sentences into atomic statements and employing embedding models to compute similarity values. The idea behind the scene is is the use of \textbf{atomic facts}. We decompose the original sentence below as an example: \\
\textit{``Theron Shan is a man who has given over his life in service to the Republic, using work to try and cope with abandonment issues gained from being hurt too many times by those who were supposed to love him.''} can be separated into:
\begin{itemize}
    \item Theron Shan is a man.
    \item He has devoted his life to serving the Republic.
    \item He uses work to cope with abandonment issues.
    \item These abandonment issues stem from being repeatedly hurt by those who were supposed to love him.
\end{itemize}
\noindent The definition of atomic facts states that they are the smallest units of information that can stand alone and be evaluated independently.\cite{b18} By segmenting a passage into distinct atomic facts, we can better understand its central meaning. Particularly in question answering and RAG evaluation, methodologies based on atomic facts have achieved significant success.\cite{b16}\cite{b18}\cite{b19}\\ 
Although atomic facts have been proven highly promising for evaluation, they still face challenges when dealing with complex contexts or long contexts.\cite{b19} Researchers have therefore turned their focus to knowledge graphs, attempting to use graph algorithms to structure and operationalize resources and improve the results. For example, Yan et al. proved that knowledge graph is useful for Atomic Fact Decomposition-based problem.\cite{b38} Li et al. also propose KELDaR framework to enhance atomic facts-based ability by knowledge graph.\cite{b37}

\section{Environment Setup} 
In this section, we will discuss in detail the specific implementation steps of our experiment environment. 

\subsection{Datasets}
All experiments in this work were carried out using the datasets below:
\begin{itemize}
    \item \texttt{qinchuanhui/}\texttt{UDA-QA}\footnote{https://huggingface.co/datasets/qinchuanhui/UDA-QA}: An English question answering dataset built on Wikipedia.  Here we only take the test part.
    \item \texttt{microsoft/}\texttt{ms\_marco}\footnote{https://huggingface.co/datasets/microsoft/ms\_marco}: A question answering dataset featuring 100,000 real Bing questions and a human generated answer. 
\end{itemize}

\subsection{Baseline System Implementation}
We constructed a basic RAG system following standard design \footnote{https://huggingface.co/learn/cookbook/en/advanced\_rag}:\\
\textbf{Pre-processing}
First, we retrieve the content corresponding to all passage URLs in the dataset. The retrieved content is then stored as textual documents for subsequent processing. We then segment them into smaller chunks and utilize the \texttt{all-MiniLM-L6-v2} model \footnote{https://huggingface.co/
sentence-transformers/all-MiniLM-L6-v2} to encode them into vector representations. The resulting embeddings are then stored in a vector database.\\
\textbf{Retriever}
For each user query, we first embed it and then compute the cosine similarity to retrieve the Top-K most relevant documents from the vector database.\\
\textbf{Generator}
We use OpenAI's \texttt{GPT-4o-mini} \footnote{https://platform.openai.com/docs/models} as our generator (LLM). The relevant documents retrieved are combined with the user query into a structured prompt (LLM Prompt), fed into the generator to produce the final output.\\
\textbf{Ragas}
In our experiments, we use RAGAS v0.3.3 \footnote{https://pypi.org/project/ragas/0.3.3/}as the benchmark for comparison.\\
The whole implementation can be found \href{https://github.com/OscarDDD/KG-Based-RAG-Evaluation}{here}. 
\subsection{Human Annotations}

To better validate the different dimensions of the rag system, we constructed a human-annotated subset from the overall dataset. Specifically, we randomly sampled 10\% of the origial entries and ask two annotators with background in NLP to evaluate each instance along the dimensions of (i) factual correctness, (ii) context relevancy, (iii) response relevancy, and (iv) faithfulness.

\section{Methodology} 

The evaluator LLM utilized in the research below is  \texttt{GPT-4o-mini} and the embedding model utilized for semantic similarity is \texttt{all-MiniLM-L6-v2}. Building upon the RAGAS, we introduce a KG-based approach that enables deeper multi-hop reasoning. The KG-based evaluation metrics we adopt are \textbf{context-agnostic}, meaning they can be flexibly applied to various combinations of input components without being tied to a specific retrieval-generation pipeline. Specifically, the following input pairs can be evaluated: Context Relevancy, Factual Correctness, Faithfulness and Answer Relevancy. In the following, we describe the evaluation steps under the assumption that we are calculating \textbf{context relevancy}---i.e., measuring the semantic alignment between the input question and the retrieved context. \\
The whole evaluation process can be separated into three stages: we first introduces the construction of the knowledge graph (Section ~\ref{sec:kg}), and then presents two algorithms implemented on top of the KG (Sections ~\ref{sec:multi} and ~\ref{sec:community}).
\subsection{KG Construction} 
\label{sec:kg}
\begin{algorithm}[H]
\caption{Build Entity-Relation Graph with Structural and Semantic Edges}
\label{alg:build_graph} 
\KwIn{Input triplets $T_{\text{in}}$, Context triplets $T_{\text{ctx}}$, Similarity threshold $\tau$}
\KwOut{Entity-relation graph $G$}
Initialize two empty graphs $G_{\text{in}}$ and $G_{\text{ctx}}$\;
\ForEach{$(h, r, t)$ in $T_{\text{in}}$ with index $i$}{
    $h_{\text{node}} \leftarrow h\_in$;\quad $r_{\text{node}} \leftarrow r\_i\_in$;\quad $t_{\text{node}} \leftarrow t\_in$\;
    Add nodes with attributes (type, group=input, original\_label)\;
    Add edges: $h_{\text{node}} \rightarrow r_{\text{node}}$ and $r_{\text{node}} \rightarrow t_{\text{node}}$ with weight 0.9, cost 0.1\;
}
\ForEach{$(h, r, t)$ in $T_{\text{ctx}}$ with index $j$}{
    $h_{\text{node}} \leftarrow h\_ctx$;\quad $r_{\text{node}} \leftarrow r\_j\_ctx$;\quad $t_{\text{node}} \leftarrow t\_ctx$\;
    Add nodes with attributes (type, group=context, original\_label)\;
    Add edges: $h_{\text{node}} \rightarrow r_{\text{node}}$ and $r_{\text{node}} \rightarrow t_{\text{node}}$ with weight 0.9, cost 0.1\;
}
Merge $G_{\text{in}}$ and $G_{\text{ctx}}$ to obtain $G$\;
$V_{\text{in}} \leftarrow$ entity nodes in $G$ ending with \_in\;
$V_{\text{ctx}} \leftarrow$ entity nodes in $G$ ending with \_ctx\;
Compute embeddings for original labels in $V_{\text{in}}$ and $V_{\text{ctx}}$ using a sentence encoder\;
Compute cosine similarity matrix $S$\;
\ForEach{$v_i \in V_{\text{in}}$}{
    \ForEach{$v_j \in V_{\text{ctx}}$}{
        \If{$S[v_i][v_j] \geq \tau$}{
            Add edge $(v_i, v_j)$ to $G$ with relation=SIMILAR, weight=$S[v_i][v_j]$, cost=$1 - S[v_i][v_j]$\;
        }
    }
}
\Return{$G$}
\end{algorithm}
\noindent We aim to construct a global knowledge graph that includes both the input and the context, as shown in the Algorithm~\ref{alg:build_graph}.

 \begin{enumerate}
    \item We first use an LLM to extract atomic factual triplets of the form $(h, r, t)$, where:             $h$: subject (\texttt{head}), $r$: relation, $t$: object (\texttt{tail})
        
    Triplets are extracted separately for both the input and the context and used as the foundation of our KG, as illustrated in Figure~\ref{fig:KG}
    
    \item We construct two disjoint KGs: one for the input and one for the context. Each triple is transformed into a mini subgraph.
        \begin{itemize}
            \item Each subject, relation, and object is treated as a distinct node.
            \item Each triple generates two directed edges:
            \begin{itemize}
                \item From head to relation (\texttt{"H-R"})
                \item From relation to tail (\texttt{"R-T"})
            \end{itemize}
        \end{itemize}
    To ensure node uniqueness, each relation is given a unique suffix (e.g., is\_1), preventing unrelated triplets with the same label (like is) from merging incorrectly. This preserves triplet independence and avoids false links. Structural edges are assigned high-confidence weights (0.9) with low cost (0.1). All nodes also carry a suffix (\_in or \_ctx) to indicate their source for clearer visualization.
These graphs are encoded using a graph data structure implemented via \texttt{NetworkX}\footnote{https://networkx.org/}, with additional metadata associated with each node:
        \begin{itemize}
            \item \textbf{original label}: the exact name of the node (e.g., relation name or entity name)
            \item \textbf{type}: the role of the node in the triple (i.e., head, relation, or tail)
            \item \textbf{group}: indicates whether the node comes from the \texttt{input} or the \texttt{context}
        \end{itemize}

\item After constructing the initial triplet-based graphs for both the input and the context, we proceed to establish semantic links across the two graphs. This step aims to identify conceptual overlaps and soft alignments between the two sources by introducing a separate relation called \texttt{SIMILAR}. The procedure is as follows:
\begin{itemize}
    \item Extract all entity nodes (i.e., \texttt{head} and \texttt{tail} nodes) from both the input and context graphs.
    \item Encode each node's original label into a high-dimensional vector using a pre-trained sentence embedding model (e.g., Sentence-BERT).
    \item Compute pairwise cosine similarity scores between all entity pairs across the two graphs.
    \item If the similarity score exceeds a threshold $\tau$ (e.g., 0.7), a \texttt{SIMILAR} edge is added between the matched nodes.
\end{itemize}
Each added edge is assigned the following attributes:
\begin{itemize}
    \item \textbf{Edge weight:} equal to the cosine similarity score
    \item \textbf{Edge cost:} defined as $1 - \text{similarity}$
\end{itemize}

This formulation implies that higher similarity (larger weight) results in a lower cost. Since edge weights represent semantic similarity in our graph, a higher weight means that the two connected nodes are semantically closer and can be treated as near equivalents, thereby justifying a lower traversal cost. \\
These semantic edges provide critical but flexible connections between the two otherwise disjoint graphs. This structure enables downstream multi-hop graph algorithms to traverse across both sources and supports fine-grained reasoning for factual consistency evaluation.
\end{enumerate}

\subsection{Multi-Hop Semantic Matching} 
\label{sec:multi}
We formalize the \texttt{input} and \texttt{context} knowledge structures as two initially disjoint subgraphs:
\[
G_{\text{in}} = (V_{\text{in}}, E_{\text{in}}), \quad G_{\text{ctx}} = (V_{\text{ctx}}, E_{\text{ctx}})
\]
These are merged into a unified KG \( G = (V, E) \), where \( V = V_{\text{in}} \cup V_{\text{ctx}} \) and \( E = E_{\text{in}} \cup E_{\text{ctx}} \cup E_{\text{sim}} \). The set \( E_{\text{sim}} \) contains semantic cross-graph edges between original labels of entity nodes with cosine similarity above a threshold \( \tau \):
\[
E_{\text{sim}} = \left\{ (v_i, v_c) \mid v_i \in V_{\text{in}},\, v_c \in V_{\text{ctx}},\, \cos(\mathbf{e}_{v_i}, \mathbf{e}_{v_c}) > \tau \right\}
\]
Under the assumption of semantic relatedness between \texttt{input} and \texttt{context}, we expect at least one path in \( G \) to connect nodes from \( V_{\text{in}} \) to \( V_{\text{ctx}} \). The original task is then transformed into one graph path search challenge:
\[
\exists\, v_i \in V_{\text{in}},\, v_c \in V_{\text{ctx}} \text{ such that } \text{cost-path}(v_i, v_c) \leq \delta
\]
where \( \delta \) is a cost threshold for traversability.
As explained in Algorithm~\ref{alg:multi-hop} and Figure~\ref{fig:multi-hop&semantic}:
\begin{enumerate}
    \item We apply a weighted version of Dijkstra's algorithm to search, for each input node, whether there exists a path to at least one context node at the given cost. The given cost serves as an effective way to avoid the issue of reaching a context node through a chain of weakly similar nodes.\cite{b30}
    \item Finally, we calculate the score based on the Formula~\ref{eq:path_score}:
\noindent
\begin{equation}
\text{Score}(G) = \frac{|\{v \in V_{\text{in}} \mid \exists\ \text{semantic path from } v \text{ to some } u \in V_{\text{ctx}} \}|}{|V_{\text{in}}|}
\label{eq:path_score}
\end{equation}
\end{enumerate}

\begin{algorithm}[H]
\caption{Multi-Hop Semantic Matching}
\label{alg:multi-hop}
\KwIn{Graph $G$, Cost threshold $\delta$}
\KwOut{Proportion of input nodes that can reach any context node}
$V_{\text{in}} \leftarrow$ nodes ending with \_in and type $\in$ \{head, tail\}\;
$V_{\text{ctx}} \leftarrow$ nodes ending with \_ctx and type $\in$ \{head, tail\}\;
\If{$V_{\text{in}} = \emptyset$ or $V_{\text{ctx}} = \emptyset$}{
    \Return 0.0\;
}
$m \leftarrow 0$\;
\ForEach{$v \in V_{\text{in}}$}{
    Compute shortest path lengths $L$ from $v$ using Dijkstra with edge cost\;
    \If{there exists $u \in V_{\text{ctx}}$ such that $L[u] \leq \delta$}{
        $m \leftarrow m + 1$\;
    }
}
\Return{$m / |V_{\text{in}}|$}
\end{algorithm}

\subsection{Community-Based Semantic Overlap}
\label{sec:community}
\begin{figure}[H]
  \centering
  \begin{subfigure}{0.9\textwidth}
    \centering
    \includegraphics[width=\linewidth, keepaspectratio]{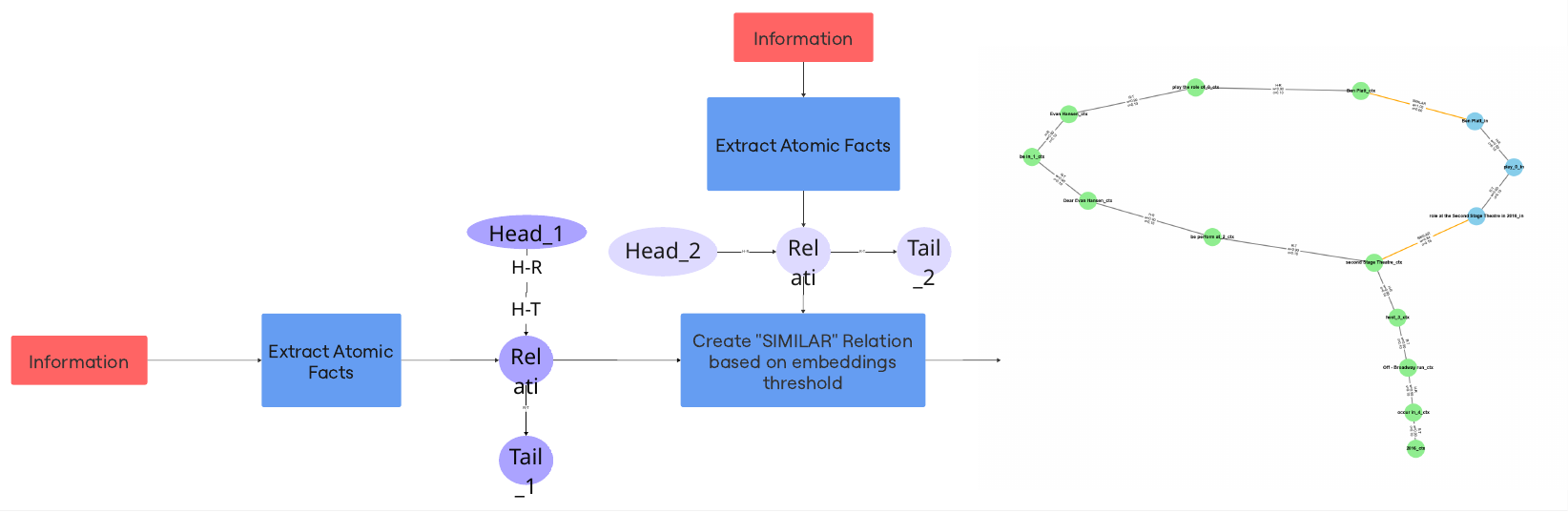}
    \caption{KG Construction}
    \label{fig:KG}
  \end{subfigure}

  \vskip 1em

  \begin{subfigure}{0.9\textwidth}
    \centering
    \includegraphics[width=\linewidth]{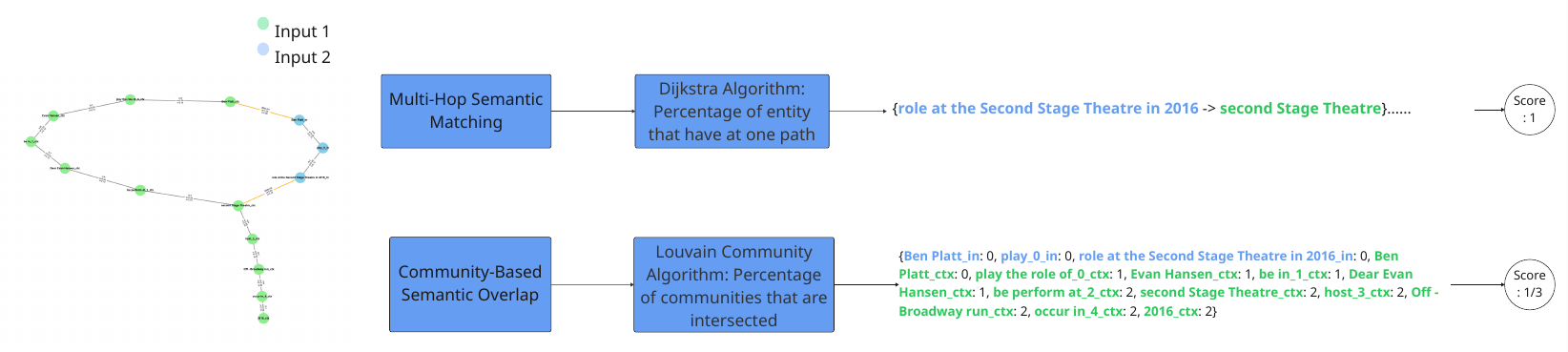}
    \caption{Multi-Hop Semantic Matching and Community-Based Semantic Overlap}
    \label{fig:multi-hop&semantic}
  \end{subfigure}

  \caption{Comparison of KG Construction and Multi-Hop Semantic Matching}
  \label{fig:two_vertical}
\end{figure}

As illustrated in Algorithm~\ref{alg:community-based} and Figure~\ref{fig:multi-hop&semantic}, the core idea of this method is that if the \texttt{input} and \texttt{context} are semantically similar, their nodes are more likely to be grouped into the same communities.

\begin{enumerate}
    \item We apply the Louvain community detection algorithm on the combined KG constructed earlier. This method partitions the graph into communities based on modularity optimization.\cite{b31}
    
    \item We then compute the final
    score using the Formula~\ref{eq:community_score}:
\noindent
\begin{equation}
\text{Score}(G) = \frac{1}{|V_{\text{in}}|} \sum_{v \in V_{\text{in}}} \mathbb{1}\left( \exists u \in V_{\text{ctx}} \text{ such that } C(v) = C(u) \right)
\label{eq:community_score}
\end{equation}
\end{enumerate}
\begin{algorithm}[H]
\caption{Community-Based Semantic Overlap}
\label{alg:community-based}
\KwIn{Graph $G$}
\KwOut{Proportion of communities covering both input and context entities}
Compute Louvain partition $P$ on $G$\;
Group nodes into communities $C$ using $P$\;
$m \leftarrow 0$\;
\ForEach{community $c \in C$}{
    $H \leftarrow$ nodes in $c$ ending with \_in and type $\in$ \{head, tail\}\;
    $T \leftarrow$ nodes in $c$ ending with \_ctx and type $\in$ \{head, tail\}\;
    \If{$H \neq \emptyset$ and $T \neq \emptyset$}{
        $m \leftarrow m + 1$\;
    }
}
\Return{$m / |C|$}
\end{algorithm}

\section{Result}
\subsection{Empirical Evaluation}
\begin{figure}[htbp]
  \centering
  
  \begin{subfigure}[t]{0.48\textwidth}
    \centering
    \includegraphics[width=\linewidth]{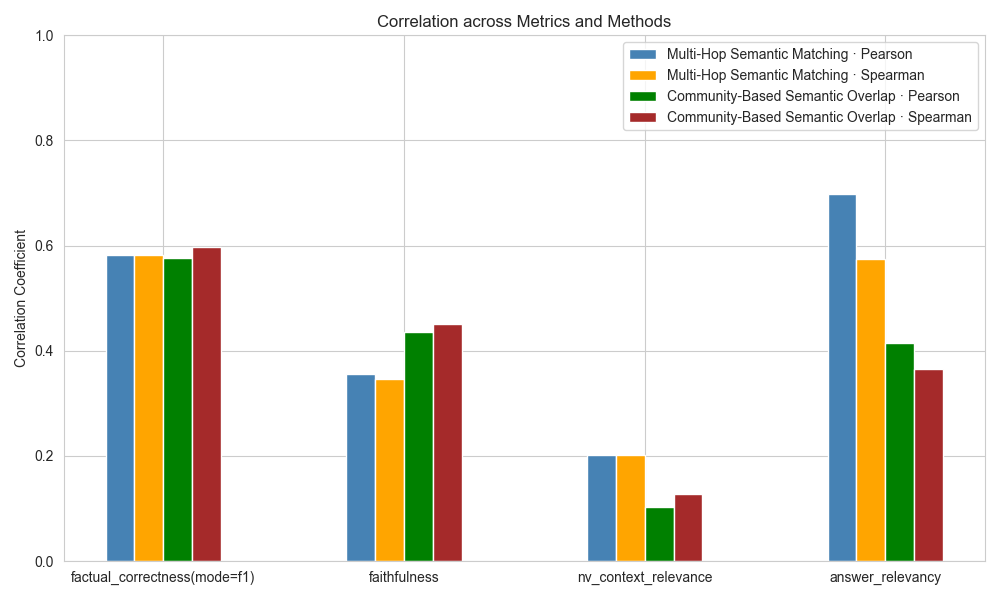}
    \caption{Correlation KG vs RAGAS -- dataset UDA-QA}
    \label{fig:kg vs ragas UDA-QA}
  \end{subfigure}
  \hfill
  \begin{subfigure}[t]{0.48\textwidth}
    \centering
    \includegraphics[width=\linewidth]{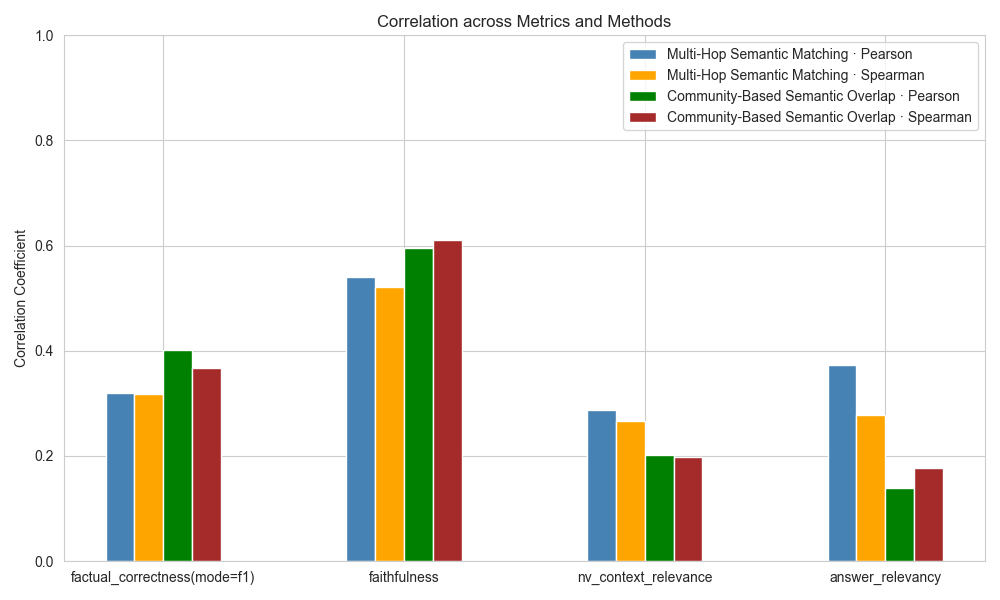}
    \caption{Correlation KG vs RAGAS -- dataset ms\_macro}
    \label{fig:kg vs ragas ms}
  \end{subfigure}
  
  \vskip\baselineskip
  \begin{subfigure}[t]{0.98\textwidth}
    \centering
    \includegraphics[width=\linewidth]{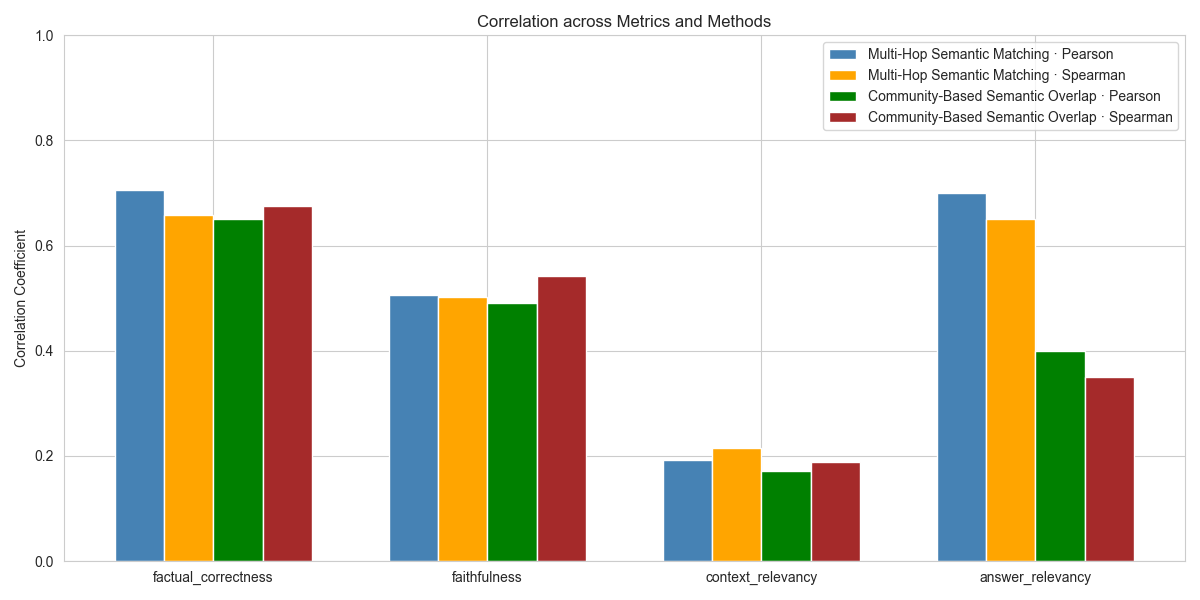}
    \caption{Correlation with Human Annotation}
    \label{fig:human}
  \end{subfigure}
  
  \vskip\baselineskip
  \begin{subfigure}[t]{0.48\textwidth}
    \centering
    \includegraphics[width=\linewidth]{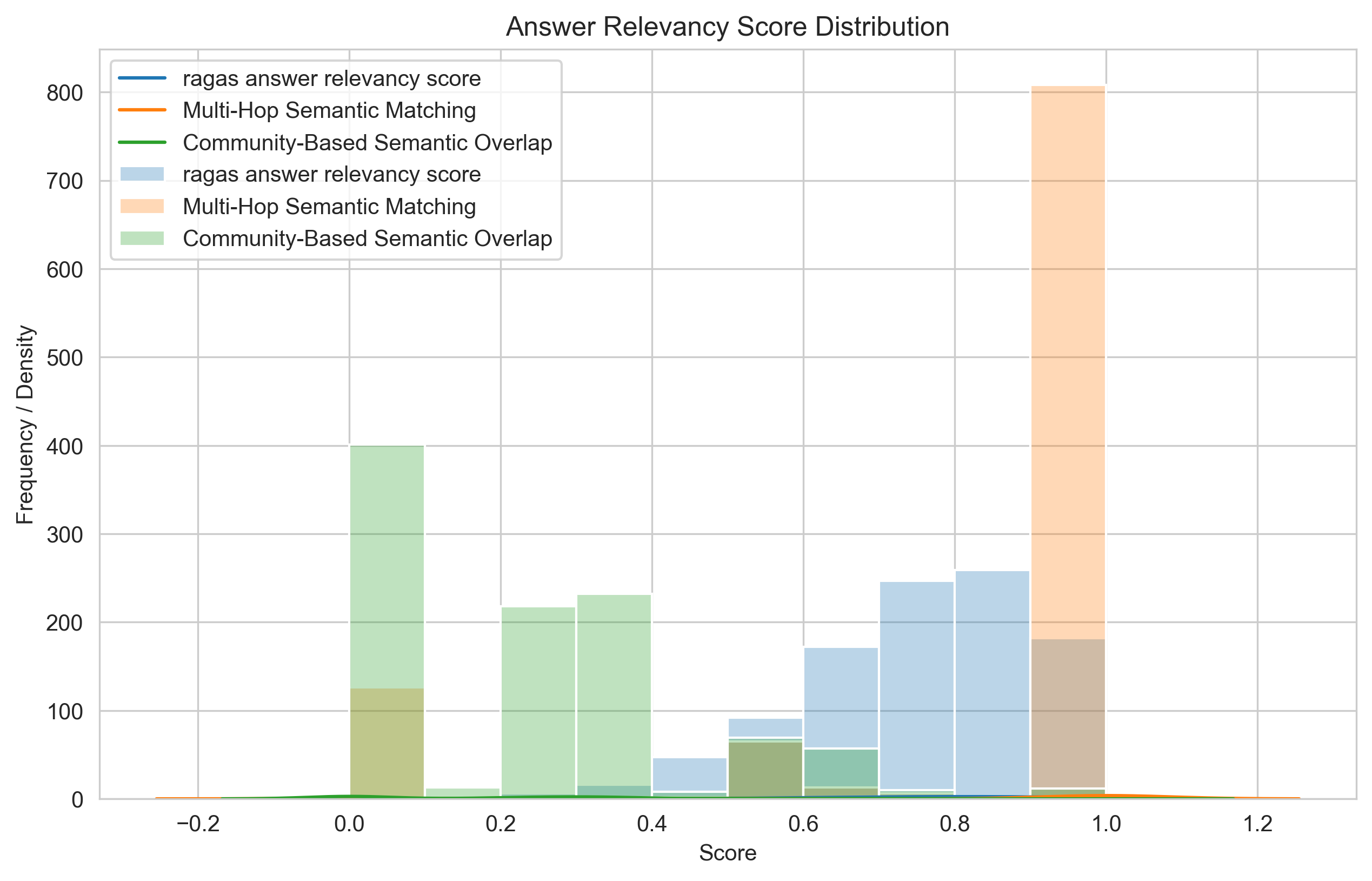}
    \caption{Ref. Answers as Substitutes -- UDA-QA}
    \label{fig:ref as answer UDA}
  \end{subfigure}
  \hfill
  \begin{subfigure}[t]{0.48\textwidth}
    \centering
    \includegraphics[width=\linewidth]{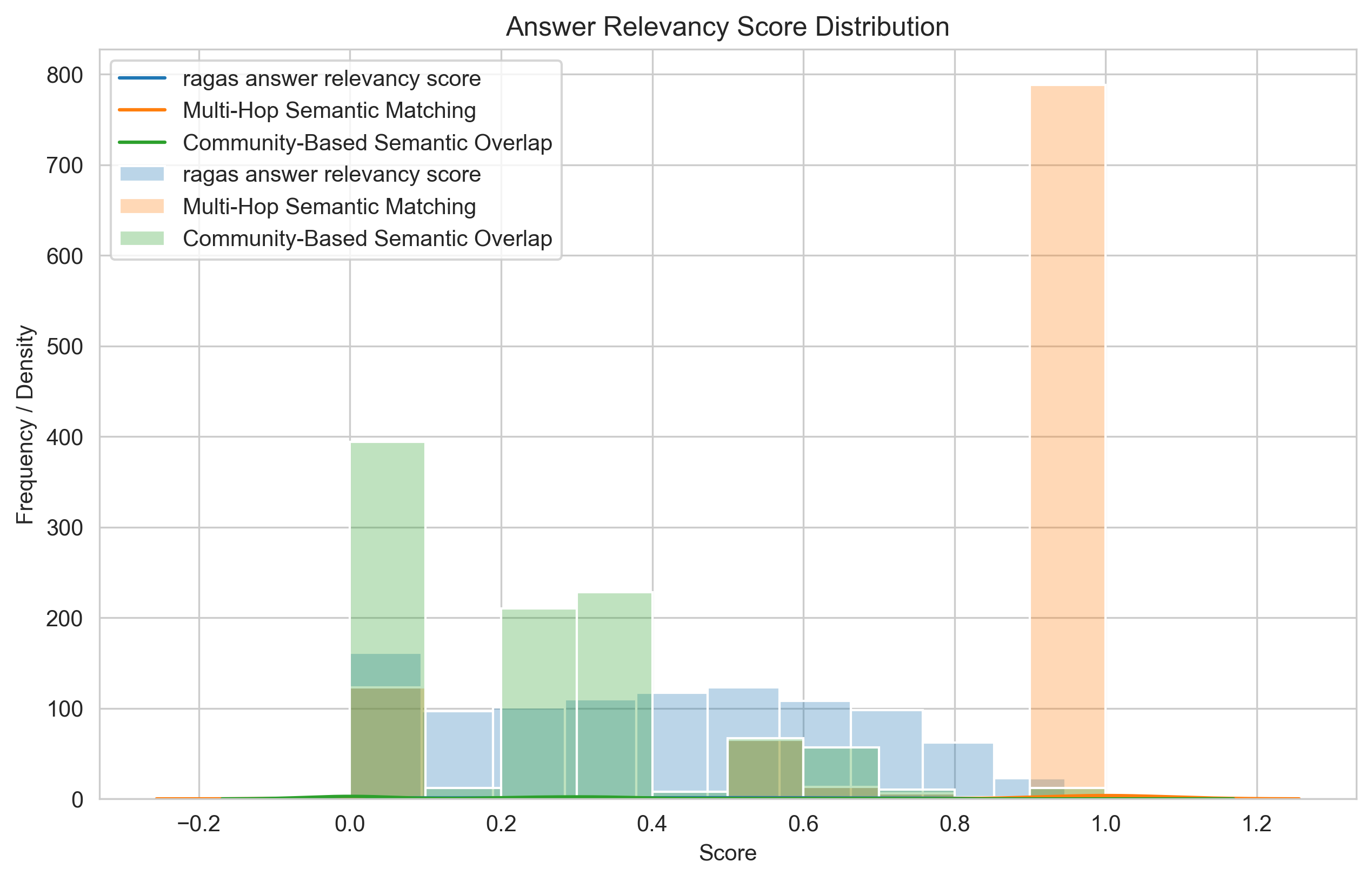}
    \caption{Ref. Answers as Substitutes -- ms\_macro}
    \label{fig:ref as answer ms}
  \end{subfigure}

  \vskip\baselineskip
  \begin{subfigure}[t]{0.48\textwidth}
    \centering
    \includegraphics[width=\linewidth]{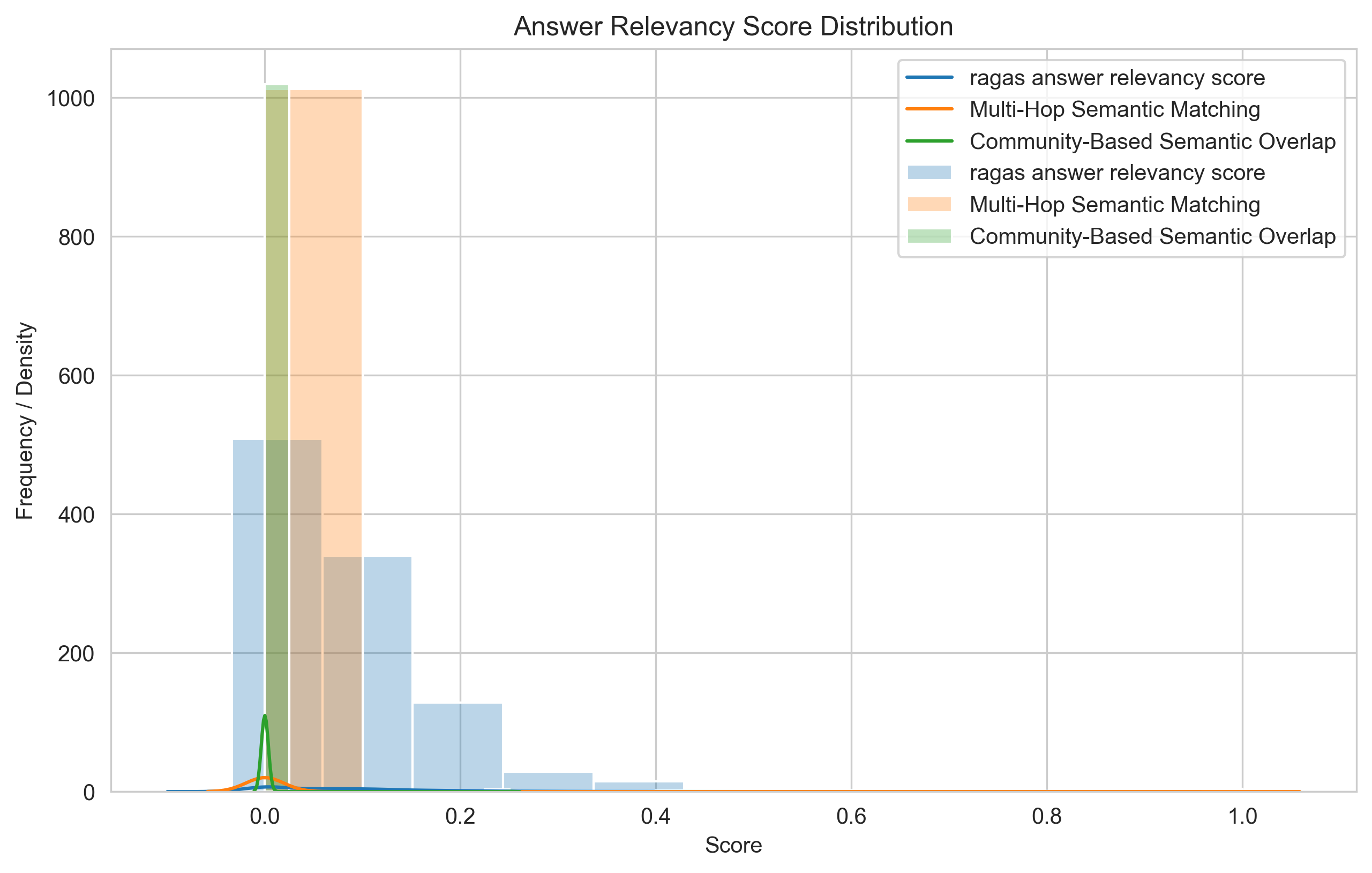}
    \caption{Wrong Answers as Substitutes -- UDA-QA}
    \label{fig:wrong answer UDA}
  \end{subfigure}
  \hfill
  \begin{subfigure}[t]{0.48\textwidth}
    \centering
    \includegraphics[width=\linewidth]{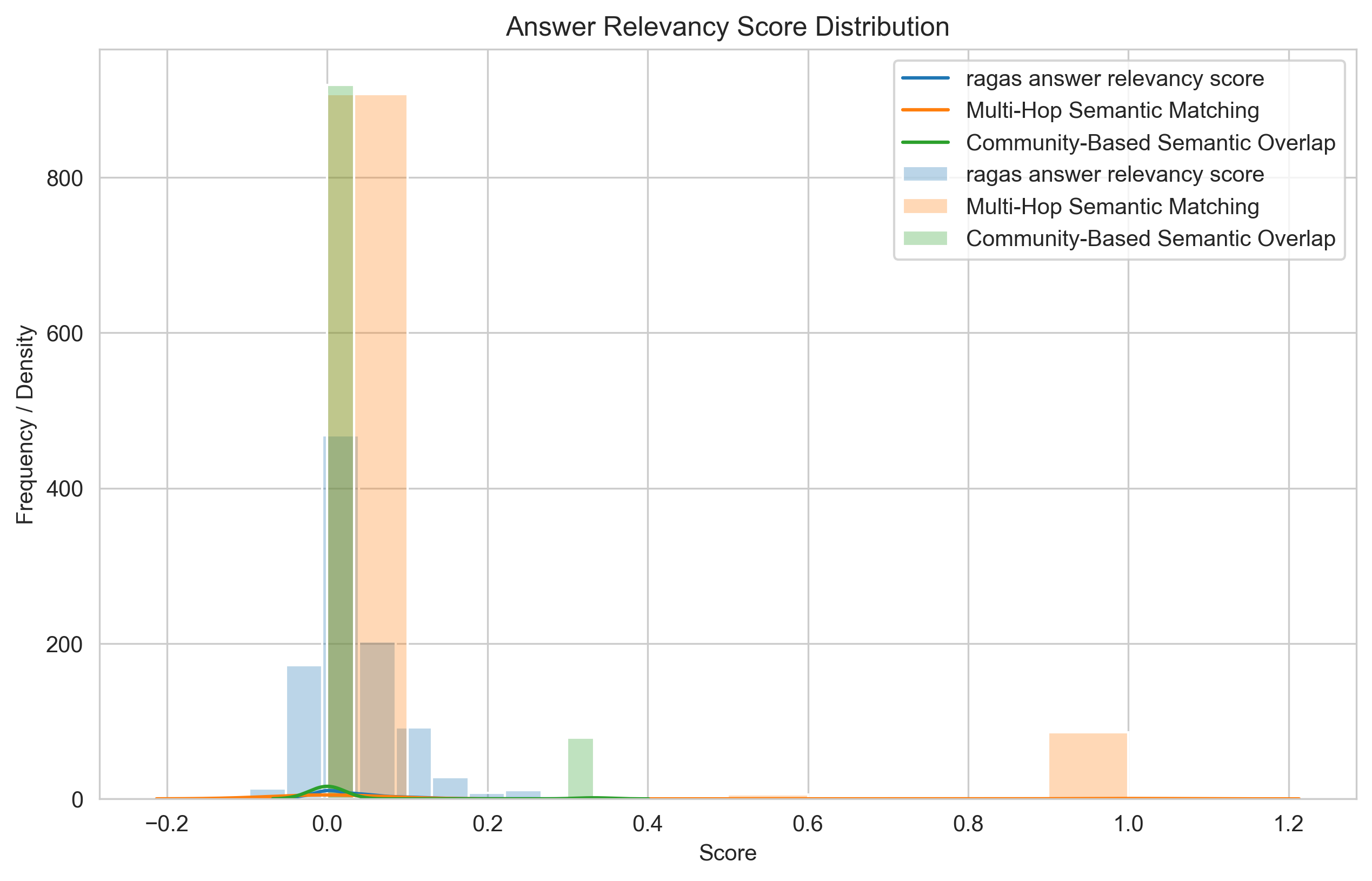}
    \caption{Wrong Answers as Substitutes -- ms\_macro}
    \label{fig:wrong answer ms}
  \end{subfigure}

 \caption{Comprehensive overview of all experimental results across datasets.}
  \label{fig:all_7}
\end{figure}

This section presents the empirical evaluation of our proposed KG-based evaluation methods with RAGAS and human annotation by assessing the correlation between them. Additionally, we analyze the sensitivity of our KG-based evaluation framework. All the results below are under the assumption that the cost is 0.5 and the threshold is 0.7.\\
As shown in Figure~\ref{fig:kg vs ragas UDA-QA} and Figure~\ref{fig:kg vs ragas ms}, except for the relatively low correlation in context relevancy, the KG-based metrics and RAGAS show moderate to high correlations on the other metrics. \\
The Multi-Hop Semantic Matching method excels in factual correctness and answer relevancy, but shows little correlation with faithfulness. In contrast, the Community-based Semantic Overlap method moderately correlates with faithfulness while performing weaker on the other metrics. These findings suggest the two methods are complementary: Multi-Hop is more effective for closely related entities, whereas Community-based is better suited for complex entity relationships.\\
To better demonstrate the correctness of our method, we conducted additional experiments on the human-annotated subset, comparing the correlation of our method with human annotations. As shown in Figure~\ref{fig:human}, both methods exhibit moderate to high correlation with human annotations in terms of factual correctness, faithfulness, and answer relevancy, further validating the effectiveness of our approach, though context relevancy still remains comparatively weaker.

\subsection{Sensitivity Analysis with Controlled Experiments}
To further investigate the performance differences between KG-based methods and the RAGAS benchmark, we conducted two additional controlled experiments. In these settings, we replaced the generated answers with either ground-truth reference answers or deliberately incorrect ones, as shown in Figure~\ref{fig:ref as answer UDA}, ~\ref{fig:ref as answer ms}, ~\ref{fig:wrong answer UDA} and ~\ref{fig:wrong answer ms} The underlying rationale 
\vspace{-2.3em}
\noindent is straightforward: since the question and its reference answer are expected to be semantically aligned, a reliable evaluation method should assign high relevance scores to such pairs. Conversely, it should assign low scores to incorrect answers that deviate from the question's intent. Although the RAGAS method generally assigns higher scores to reference answers and lower scores to incorrect answers, our proposed KG-based methods—particularly the \textbf{Multi-Hop Semantic Matching} approach, which produces scores that are consistently close to 1 for reference answers and nearly 0 for incorrect ones. While the \textbf{Community-Based Semantic Overlap} method performs poorly on reference answers, it demonstrates strong discriminative ability in identifying incorrect answers.

\subsection{Complementary Strengths of RAGAS and KG Methods}
The KG-based evaluation framework demonstrates an overall moderate to high correlation with RAGAS as well as the human annotation, indicating that it captures similar underlying evaluation patterns. Yet it presents a high correlation in the Answer Relevancy metric, while the Context Relevancy metric shows a significantly lower correlation.\\
This discrepancy in correlation might be attributed to the underlying principles of our algorithm. Our KG-based algorithm, especially the \textbf{Multi-Hop Semantic Matching} method, emphasizes identifying high entity-level relevance between the two inputs. Since answers often contain fewer irrelevant entities and maintain more substantial alignment with the question's entity scope, the KG methods tend to assign higher scores in these cases. On the other hand, retrieved-context typically includes a broader range of information, resulting in dispersed subgraphs with weaker connectivity and less community overlap, which lowers the scores.\\
According to the sensitivity experiments, we further confirm that \textbf{Multi-Hop Semantic Matching} is more responsive when semantic relevance is either strongly present or absent. In contrast, while RAGAS assigns scores with a directional bias in both cases, it does not exhibit a sharply distinguishable shift in distribution.\\
In conclucsion, the KG-based evaluation framework provides more sensitive insights into semantic consistency, especially under conditions of high entity-level relevance or semantic contrast and thus becomes an ideal complement to the RAGAS framework.

\section{Limitations}

A major limitation of our evaluation system lies in its scalability. The core bottleneck is the high computational cost of graph construction. In particular, when the input context is large, the time required to build the graph grows significantly, which hinders efficiency and makes scaling to real-world settings challenging.

\section{Conclusion and future scope}
This paper proposes an LLM-driven KG-based approach for evaluating RAG systems. By leveraging an LLM as an evaluator and defining multi-dimensional metrics, we conduct an efficient and accurate assessment of RAG systems.We evaluate two KG-based subscores, \textbf{Multi-Hop Semantic Matching} and \textbf{Community-Based Semantic Overlap}, which show moderate-to-high correlation with both human annotations and RAGAS. They complement each other across different metrics, and exhibit higher sensitivity when contrasting highly or non-relevant inputs. Currently, we only focus on the similarity between individual entities. Valuable research directions can be to investigate how to extend the similarity to triplet level and how to find a well-defined hyperparameter to gain a more fine-grained evaluation. Other metrics, such as negative rejection and long-context accuracy, also worth thorough exploration. \cite{b27}\cite{b28}
\begin{credits}
\subsubsection{\ackname} This research was funded by the Federal Ministry of Education and Research of Germany (BMBF) as
part of the CeCaS project, FKZ: 16ME0800K.
\subsubsection{\discintname}
The authors have no competing interests to declare that are relevant to the content of this article.
\end{credits}
%
%
%
\bibliographystyle{splncs04}
\bibliography{kg_rag_evaluation}
%

\end{document}